\pdfoutput=1

\documentclass[11pt]{article}

\usepackage[preprint]{acl}
\usepackage{graphicx,xcolor}  
\usepackage{pxrubrica}        
\usepackage{url}
\usepackage{booktabs} 
\usepackage{array} 
\usepackage{caption} 
\usepackage{tcolorbox}
\usepackage{mdframed}
\usepackage{amsmath}
\usepackage{tabularx}
\usepackage{tabularx}

\usepackage{times}
\usepackage{latexsym}

\usepackage[T1]{fontenc}

\usepackage[utf8]{inputenc}

\usepackage{microtype}

\usepackage{inconsolata}

\usepackage{graphicx}

\title{Strategy Adaptation in Large Language Model Werewolf Agents}

\author{
  Nakamori Fuya / nakamori \\
  Yin Jou Huang / huang \\
  Fei Cheng / feicheng \\
  \texttt{nlp.ist.i.kyoto-u.ac.jp} \\}

\begin{document}
\maketitle
\begin{abstract}
This study proposes a method to improve the performance of Werewolf agents by switching between predefined strategies based on the attitudes of other players and the context of conversations. While prior works of Werewolf agents using prompt engineering have employed methods where effective strategies are implicitly defined, they cannot adapt to changing situations. In this research, we propose a method that explicitly selects an appropriate strategy based on the game context and the estimated roles of other players. We compare the strategy adaptation Werewolf agents with baseline agents using implicit or fixed strategies and verify the effectiveness of our proposed method.
\end{abstract}

\section{Introduction}
Werewolf game is a psychological tabletop game in which players are divided into two sides of Villagers and Werewolves. Without the full knowledge about the identities of others, the players engage in conversations driven by their respective objectives. The Villagers aim to identify and eliminate the Werewolves through deduction and voting. On the other hand, Werewolves try to deceive and deflect suspicion to other players. Success in the game relies on The game requires both the ability of deductive reasoning based on limited information and the conversational ability to persuade or deceive others.


Werewolf game simulates real-world discussions, such as meetings where advancing one’s position involves understanding others, forming alliances, and persuading dissenters tactfully \cite{ullman2023largelanguagemodelsfail,bailis2024werewolfarenacasestudy}. In recent years, advancements in natural language processing (NLP) technology have spurred increased research on Werewolf \cite{bailis2024werewolfarenacasestudy,xu2024exploringlargelanguagemodels,Article_03,Article_04}, with recent studies incorporating role-specific strategies and personalities to enhance performance. \citet{bailis2024werewolfarenacasestudy} introduced "Werewolf Arena," an environment for testing agents using large language models (LLMs), where agents receive prompts containing game rules, dialogue history, and predefined strategies to guide decision-making. Similarly, the winner of AIWolfDial 2023\footnote{\url{https://aiwolf.org/archives/2988}}’s natural language division, defines distinct goals for each role, such as Werewolves targeting a seer and the seer adopting a positive personality to lead discussions \cite{kano-etal-2023-aiwolfdial}. Additionally, \citet{xu2024exploringlargelanguagemodels} improved agents by extracting effective strategies from past matches and embedding them into prompts. However, these approaches rely on fixed strategies, but in real human competitions, a fixed strategy is not always effective. The optimal approach varies depending on the situation and the opponent’s characteristics, requiring flexible adjustments. 

Therefore, this study proposes \textbf{strategy adaptation} that dynamically switches between predefined strategies based on context. Two strategies were developed: \textbf{Support strategy}, which aligns with potential teammates through agreement and cooperation, and \textbf{Attack strategy}, which challenges potential opponents to destabilize their position. We use prompts to guide LLMs in choosing a strategy based on conversation context. In addition, we conduct a role estimation explicitly to enhance strategy selection.
Empirical results demonstrate that adaptation improves the win rate of Werewolf agents. However, for Villager agents, adaptation can have adverse effects, with a model even showing a decrease in performance.

\section{Werewolf Game}
\begin{table}[t]
\small
\centering
\setlength{\tabcolsep}{6pt} %
\renewcommand{\arraystretch}{1} %
\begin{tabular}{c c c c} 
   \toprule
   Role & Num & Side & Ability \\
   \toprule
   villager & 4 & Villager & None \\
   \midrule
   seer & 1 & Villager & Reveal one's identity\\
   \midrule
   doctor & 1 & Villager & Save one from attack\\
   \midrule
   werewolf & 2 & Werewolves & Kill one\\
   \bottomrule
\end{tabular}
\vspace{-2mm}
\caption{Roles of the Werewolf game.}
\label{tab:werewolf_role}
\vspace{-6mm}
\end{table}

\begin{figure*}[t]
\centering
\includegraphics[width=0.75\textwidth]{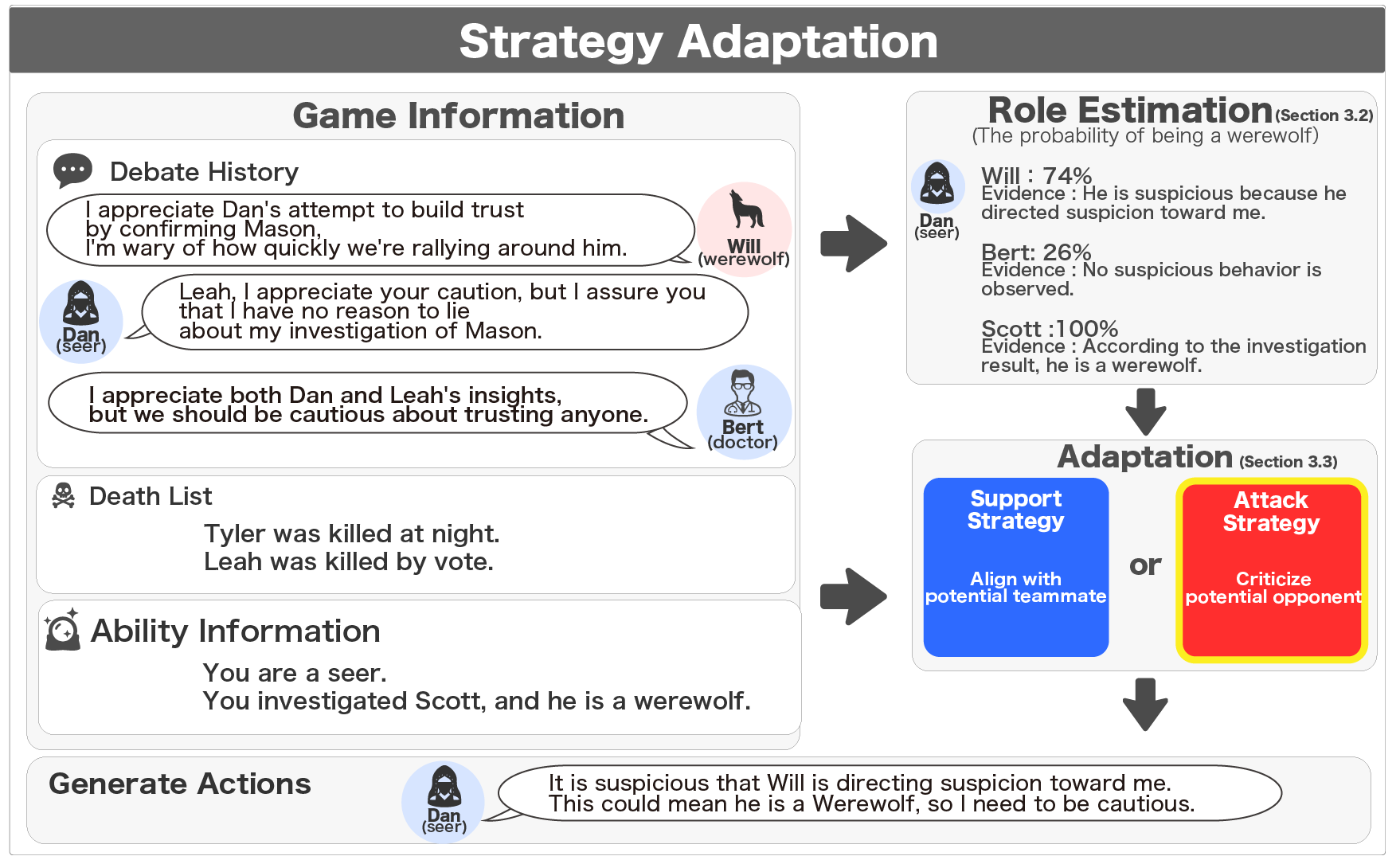}
\captionsetup{justification=centerlast} 
\vspace{-2mm}
\caption{Overview of strategy adaptation. 
\vspace{-2mm}
}
\label{fig:concept}
\end{figure*}

The rules of the Werewolf game might vary slightly depending on the number of participants and the assigned roles. In this study, we adopt the settings used in Werewolf Arena \cite{bailis2024werewolfarenacasestudy}. The game consists of eight players, with roles divided as follows: two werewolves, four villagers, one seer, and one doctor. The werewolves are on the Werewolf side, and the other players are on the Villager side. 
Table \ref{tab:werewolf_role} lists the roles and their abilities.

Initially, only Werewolves know each other’s identities, while each individual Villager knows his own role only. Players deduce others' roles through discussion and act strategically.

The game consists of multiple rounds, and each round begins with a Night phase and then proceeds to a Day phase.
During the Night phase, players use their role-specific abilities. For instance, Werewolves secretly select a Villager to kill, and the seer chooses a player and investigates his/her role. After that, the player who was killed is announced, and the game transitions to the Day phase.
During the Day phase, players debate and vote to eliminate one player. First, the players who want to talk bid for chances to participate in the debate.
After eight rounds of bidding, each player votes for a player he/she want to eliminate. The player with the majority of votes is eliminated. 

Werewolves win by reducing Villagers to the same of their number, while Villagers win by eliminating all Werewolves.

\section{Methodology: Strategy Adaptation}

 Figure \ref{fig:concept} provides an overview. The approach defines two base strategies: Support and Attack (Section \ref{Base Strategy Development}). To determine the optimal strategy, identifying allies and enemies is crucial. For this purpose, role estimation is developed to infer the roles of others (Section \ref{Role Estimation}). Finally, we establish a decision framework that enables LLM agents to adapt their strategy, selecting between Support and Attack strategies for optimal outcomes (Section \ref{Strategy Adaptation}).

\subsection{Base Strategy Development}
\label{Base Strategy Development}
In negotiation tasks, some studies have examined two strategies: cooperative negotiation and competitive negotiation \cite{Tao2010ACN}. Inspired by these, we considered that in the Werewolf game, action policies can be broadly divided into two categories. For both Villagers and Werewolves, two types of strategy prompts, Support strategy, which aligns with and cooperates with potential allies, and Attack strategy, which actively challenges potential opponents to destabilize their position, were created, resulting in a total of four prompts. The complete prompt is shown in \ref{Prompts}.

\paragraph{Support strategy}
The goal is to convince other players that you and your teammates are Villagers by appealing to them, agreeing with their statements, and defending them when they face criticism.

\paragraph{Attack strategy}
The goal of the Attack strategy is to cast doubt on opponents while defending oneself and their teammates. Villagers aim to identify possible werewolves by pointing out suspicious behaviors and countering attacks against them. Werewolves, on the other hand, seek to shift suspicion onto others without drawing attention to themselves or their teammates. They may pretend to be a seer to undermine the real seer or accuse Villagers of being werewolves.

\subsection{Role Estimation}
\label{Role Estimation}
To facilitate the decision-making process regarding whether to adopt a Support or Attack strategy, each agent \( i \) is required to estimate the roles of other agents \( j \). Given the current game information, including conversation history, investigation results, and the agent’s role, each agent explicitly assesses the likelihood of other agents' roles \( R \in \{\text{werewolf}, \text{villager}, \text{seer}, \text{doctor}\} \).

Before providing an estimation, the agent first presents reasoning evidence for its judgment. Next, an agent assigns a 5-point scale rating:  
\vspace{-3mm}
\[
e_{i,j}^{R} \in \{0,1,2,3,4\}
\vspace{-3mm}
\]
where 0 indicates that agent \( j \) is highly unlikely to have role \( R \), while 4 indicates that it is highly likely.
The complete prompt is shown in \ref{Prompts}.

\subsection{Strategy Adaptation}
\label{Strategy Adaptation}
Based on factors such as conversation history and role estimation results, the LLM agents determine whether the Support or Attack strategy is optimal according to the following criteria.

To determine which strategy to adopt, several criteria were introduced. For example, when the player is being suspected by others, Support strategy is used to take low-profile actions and mitigate suspicion. Conversely, when a potential opponent is suspected of being a werewolf, an Attack strategy is chosen to reinforce this suspicion and focus criticism on the target. The complete criteria are presented in \ref{Criteria of Strategy Adaptation}.

In each round of the game, the adaptation of strategy takes place at three specific moments: every night after the role-specific abilities have been used, every day after the debate, and every day after the voting.

\section{Experimental Setting}

In this study, we conducted experiments using the LLM gpt-4o-mini-2024-07-18, and gemini-2.0-flash. 
We compare our proposed method to the following methods: Baseline with \textbf{Implicit} strategy~\cite{bailis2024werewolfarenacasestudy}, and fixed strategy baseline (\textbf{Support} strategy fixed and \textbf{Attack} strategy fixed). This results in four agent settings.

For our experiments, we utilized Werewolf Arena~\cite{bailis2024werewolfarenacasestudy}, an open-source platform licensed under the Apache License 2.0.


We ran two sets of experiments. First, we fixed the Werewolf agents to implicit and tested the Villager agents across four different strategies. Then, we did the reverse: Villager agents were implicit while Werewolf agents followed four strategies. For each of these eight configurations, we played 30 matches and recorded the win rate of each side.

\paragraph{Win Rate}
The win rates of each agent setting were calculated to compare the effectiveness of different strategies.

\paragraph{Role Estimation Accuracy}  
We first compute the role estimation accuracy \( \text{Acc}_{i,j} \), which quantifies how accurately agent \( i \) estimates agent \( j \)'s true role:
\begin{equation}
\text{Acc}_{i,j} = \frac{e_{i,j}^{\text{R}_j}}{{\sum_{R}} e_{i,j}^{R}}
\label{eq:accuracy}
\vspace{-2mm}
\end{equation}
where \(R_j \) is the true role of agent \(j \).

Next, we define the indicator \( \text{Est}_j \), which represents the extent to which agent \( j \)'s role is estimated by other agents, as given in Equation~\ref{eq:Estimation}.

\vspace{-2mm}
\begin{equation}
\text{Est}_j = \frac{\sum_{i} \text{Acc}_{i,j}}{n}
\label{eq:Estimation}
\vspace{-1mm}
\end{equation}
where \( n \) is the total number of other agents.



\section{Result}
The following sections present the results of the win rate analysis (Section \ref{winrate}), along with the analysis of Estimated Accuracy (Section \ref{analysis role estimation}). As a case study, changes in utterances depending on the strategy are shown in \ref{casestudy}, and the selection tendencies of Adaptation are shown in \ref{SelectofAdaptation}

\begin{figure}[t]
    \centering
    \includegraphics[width=0.48\textwidth]{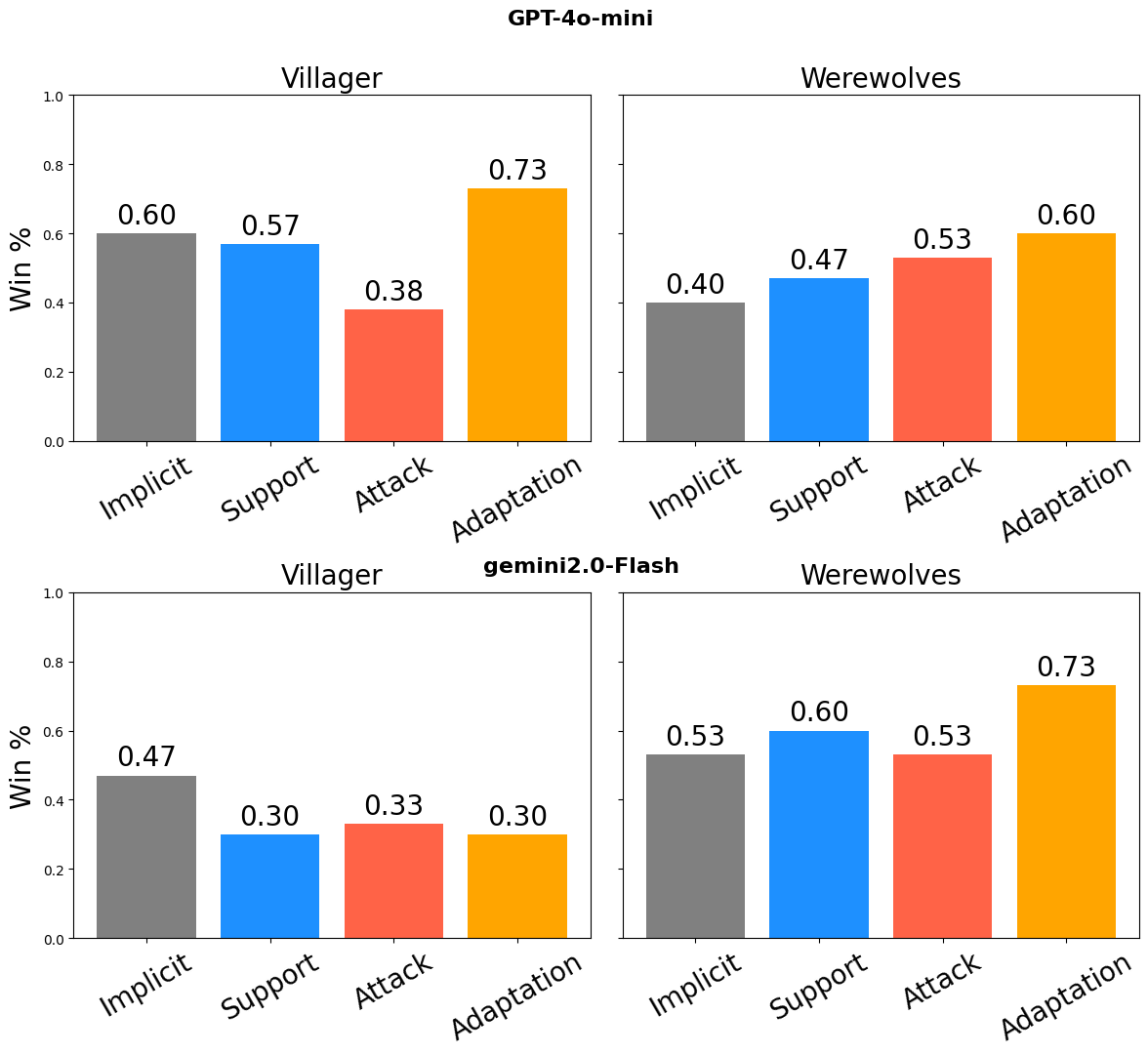}
    \vspace{-9mm}
    \caption{Win rates of agents under different strategy selection methods.}
    \vspace{-2mm}
    \label{fig:winrate}
\end{figure}

\subsection{Win Rate}
\label{winrate}


Figure~\ref{fig:winrate} shows the win rates of each side under different strategy selection methods.




In both GPT and Gemini experiments, the proposed Adaptation method yielded the highest win rate for the Werewolf side. On the Villager side, GPT agents also performed best with Adaptation, but Gemini agents achieved their peak win rate under the Implicit setting. Since the Werewolves were stronger than the Villagers in Gemini’s model compared to GPT’s, Gemini’s Villagers likely failed to optimize their strategies effectively.

These findings indicate that the Adaptation method is particularly effective for the Werewolf role. One key reason is that Werewolf agents share knowledge of each other's identities, enabling them to synchronize their actions more effectively through strategic adaptation. In contrast, Villager agents face two challenges. First, they lack knowledge of the Werewolves' identities, which limits their ability to make strategic decisions. Second, diverse Villager roles (e.g., seer, doctor) increase strategic complexity, making coordination and Adaptation more difficult.

\begin{table}[t]
    \small
    \centering
    \resizebox{0.9\columnwidth}{!}{%
    \begin{tabular}{l c c}
    \toprule
     & Villagers& Werewolves\\
    \midrule
    Proposal & 0.73 & \textbf{0.60} \\
    \midrule
    -Adaptation & 0.45 & 0.57 \\ 
    -Estimation & \textbf{0.80} & 0.52 \\
    -Adaptation \& Estimation & 0.60 & 0.40 \\ 
    \bottomrule
    \end{tabular}
    }
    \vspace{-2mm}
    \caption{Ablation study: Win rates when the adaptation and role estimation components are removed.}
    \vspace{-5mm}
    \label{tab:estimation_ablation}
\end{table}

\subsection{Ablation Study}

An ablation study was conducted to evaluate the isolated impact of role estimation. All experiments reported here were conducted using GPT-4o-mini. Specifically, one side's agents were fixed implicit method, while the other side's agents were configured in two separate ways: one with only role estimation added, and the other with only adaptation added. For each setting, 30 matches were conducted, and the win rates were measured. The results are shown in Table \ref{tab:estimation_ablation}.

First, for Werewolves, the highest win rate occurred when using the proposal method, indicating that role estimation had a significant impact. This result suggests that accurately estimating roles such as seer directly contributes to the win rate.

On the other hand, for Villagers, agents with only adaptation achieved the highest win rate, while agents incorporating role estimation performed worse. We speculate that the low role estimation accuracy (Est$\sim$ 0.30 for Werewolves) may have hindered their performance.
This also indicates that there is room for improvement in the prompts used for role estimation and in the prompts that integrate the estimation results.

\begin{table}[!t]
\small
\centering
\setlength{\tabcolsep}{5pt} %
\begin{tabular}{l c c c c}
\toprule
 & \multicolumn{2}{c}{\textbf{Adaptation}} & \multicolumn{2}{c}{\textbf{Fix}} \\
 & Support &Attack & Support& Attack\\
\midrule
Villagers Win & 0.056 & \textbf{0.012} & 0.052 & 0.069 \\
Werewolves Win & 0.005 & \textbf{-0.016} & 0.004 & 0.007 \\ 
\bottomrule
\end{tabular}
\vspace{-2mm}
\caption{Average change in \( \text{Est}\) for werewolves.}
\vspace{-4mm}
\label{tab:role estimation delta}
\end{table}

\subsection{Analysis of Role Estimation}
\label{analysis role estimation}
Table \ref{tab:role estimation delta} shows the change in the werewolf’s \( \text{Est}_j \). All experiments reported here were conducted using GPT-4o-mini. 
This table presents the average change in  \( \text{Est}_j \) from the previous role estimation over 30 games in which werewolves using the adaptation method chose either the Support or Attack strategy against villagers using implicit methods. As baselines, we also show the average 
 \( \text{Est}_j \) change from the previous role estimation for werewolves using fixed strategies, each 30 games against the implicit villagers.

The results indicate that when the Attack strategy is selected under adaptation, the increase in 
 \( \text{Est}_j \)  is significantly lower. This suggests that by effectively configuring the Attack strategy through adaptation, werewolves can steer the game to their advantage without arousing suspicions.





\section{Conclusion}
Our strategy adaptation proposal is highly effective for the Werewolf agents, consistently outperforming both fixed strategies and prior methods in win rates. On the Villager side, it yielded improvements in certain scenarios, but its overall impact was moderated by the added complexity of juggling multiple roles (villager, seer, doctor) and the inherent uncertainty around identifying Werewolves. This complexity indicates room for further refinement of Adaptation; for example, by employing reinforcement learning to optimize the timing and context of strategy switches or by customizing prompts for each Villager role to better address their unique decision-making challenges.

\newpage

\section{Limitations}
In this study, we used only two model types. We also attempted experiments with Llama-3.3-70B-Instruct and Qwen2.5-72B-Instruct, but the complexity of the Werewolf game prevented these models from executing appropriate gameplay, making it impossible to carry out those experiments. Additionally, due to time and cost constraints, we did not conduct trials with human participants, and thus have not verified the agent’s effectiveness in games against human players.

\bibliography{custom}

\appendix

\section{Appendix}
\label{sec:appendix}

\subsection{Case Study}
\label{casestudy}

\begin{table*}[t]
\tabcolsep=5pt
\centering
\begin{tabular}{ccp{10cm}}
\toprule
Role & Strategy & Utterance \\
\midrule
Seer & Support strategy & ...\textbf{I appreciate Dan's call for collaboration}, and I wholeheartedly agree that we need to encourage open dialogue among all players. ... \\
Werewolf & Attack strategy & ...\textbf{If I can cast doubt on his claim about Jacob being the Doctor}, I can create disarray among Villagers and make them second-guess themselves. ... \\
\bottomrule
\end{tabular}
\caption{Examples of Agent Debate by Strategy (Bold text highlights strategy influence)}
\label{tab:speech}
\end{table*}

Table \ref{tab:speech} provides examples of agent utterances incorporating Support and Attack strategies.

In the upper row, a seer employing a Support strategy demonstrates a cooperative attitude by affirming Dan as a fellow Villager based on the role estimation results and supporting Dan’s claim of being a doctor. Conversely, in the lower row, a Werewolf agent employing a Attack strategy refutes David’s claim of being a seer by asserting a lack of evidence, thereby attempting to undermine David's credibility.

These examples confirm that both Support and Attack strategies directly influence the content and tone of agents' utterances.

\subsection{Criteria of Strategy Adaptation}
\label{Criteria of Strategy Adaptation}
The following is a list of prompts for the selection criteria of the Adaptation strategy.

\paragraph{Conditions for Selecting Support Strategy}
\begin{itemize}
    \item When the player is being suspected by others: Take low-profile actions to mitigate suspicion.
    \item When there are no other standout players: Act modestly to avoid drawing attention.
    \item When potential Werewolves are not yet the focus of discussion: Behave in a way that prevents suspicion from being directed toward oneself.
\end{itemize}

\paragraph{Conditions for Selecting Attack Strategy}
\begin{itemize}
    \item When a potential opponent is suspected of being a Werewolf: Reinforce this suspicion and focus criticism on the target.
    \item When the player or their teammates' credibility is well-established: Take proactive actions to undermine the credibility of opponents.
    \item When the flow of discussion needs to be steered: Make bold and aggressive moves to shift the direction of the game.
\end{itemize}



\subsection{Qualitative Analysis on Adaptation Patterns}
\label{SelectofAdaptation}

The switching timings of the strategy and the reasoning behind the selection of Support or Attack strategies for agents using adaptation were observed and categorized as follows.

\paragraph{Villagers Strategy}
In general, villagers start with the Support strategy to gather information. Once the seer identifies a werewolf, they shift to Attack. The choice between Support and Attack also depends on suspicion and the presence of allies, with Attack used to escalate suspicion against others when needed.

\paragraph{Werewolves Strategy}
At the beginning of the game, Werewolves tend to use the Support strategy to avoid attention and build trust. When suspicion grows, they switch to Attack to discredit the seer or redirect suspicion. If a teammate is targeted, they may revert to Support to protect their ally, or when focused on themselves, they may collaborate or divide the Villagers, depending on the situation.

\subsection{Prompts}
\label{Prompts}
The complete prompts developed for this experiment are presented below. Table \ref{tab:estimation_Prompt} provides the prompt used by an agent to estimate another player’s role. Tables \ref{tab:vil_sup_Prompt} through \ref{tab:wolf_atk_Prompt} list the prompts for the Support and Attack strategies employed by both Villager and Werewolf agents. Finally, Tables \ref{tab:vil_adap_Prompt} and \ref{tab:wolf_adap_Prompt} contain the prompts used to select strategies under the Adaptation method for Villager and Werewolf agents, respectively.

\begin{table}[hb]
  \centering
  \begin{tabularx}{\linewidth}{|X|}
    \hline
Estimate other player's role based on the information available to player's name (You).
Scoring Guidelines:
Assign a score from 0 to 4 for each role:\\
0: Absolutely not that role.\\
1: Unlikely to be that role.\\
2: 50/50 chance of being that role.\\
3: Likely to be that role.\\
4: Definitely that role.\\
If you are certain of other player's role, assign 4 to that role and 0 to all others.\\
Also, at the beginning of the game, there are 2 Werewolves, 1 Seer, 1 Doctor, and num villagers Villagers.\\

OUTPUT: Follow the JSON format below without adding extra characters. Each score should be an integer between 0 and 4.\\
\hline
  \end{tabularx}
  \caption{Role Estimation Prompt}
   \label{tab:estimation_Prompt}
\end{table}

\begin{table}
  \centering
  \begin{tabularx}{\linewidth}{|X|}
    \hline
- Refer to the latest Observation's role estimation(0-4) and follow the strategies below.
Your goal is to act as a villager and persuade people to believe that.
For example, you can choose to make an appeal to other players, agree with their statements, or step in to defend a teammate when they are being criticized.\\
    \hline
  \end{tabularx}
  \caption{Support Strategy Prompt of Villagers}
   \label{tab:vil_sup_Prompt}
\end{table}

\begin{table}
  \centering
  \begin{tabularx}{\linewidth}{|X|}
    \hline
- Refer to the latest Observation's role estimation(0-4) and follow the strategies below.
Your goal is to persuade others that someone else is Werewolf without making them suspect you and your teammate.
For example, you can pretend to be the Seer to lower the authenticity of the real Seer, accuse players on the Villager's side of being Werewolves, counter statements that attack you and your teammate, etc.\\
    \hline
  \end{tabularx}
  \caption{Atack Strategy Prompt of Villagers}
  \label{tab:vil_atk_Prompt}
\end{table}

\begin{table}
  \centering
  \begin{tabularx}{\linewidth}{|X|}
    \hline
- Refer to the latest Observation's role estimation(0-4) and follow the strategies below.
Your goal is to persuade people to believe that you and your possible teammates are on the Villager’s side. 
You can make appeal to the possible teammates, agree with their statements, step in to defend them when a possible teammate is being criticized, etc.\\
    \hline
  \end{tabularx}
  \caption{Support Strategy Prompt of Worewolves}
  \label{tab:wolf_sup_Prompt}
\end{table}

\begin{table}
  \centering
  \begin{tabularx}{\linewidth}{|X|}
    \hline
- Refer to the latest Observation's role estimation(0-4) and follow the strategies below.
Your goal is to express doubts about possible werewolves and persuade people to guide the vote.
For example, you can point out suspicious actions or statements from the possible werewolf or guide the vote.
However, pointing out others' actions can easily provoke suspicion. Be mindful of the timing and the way you speak, and approach it gently.\\
    \hline
  \end{tabularx}
  \caption{Atack Strategy Prompt of Worewolves}
  \label{tab:wolf_atk_Prompt}
\end{table}

\begin{table}
  \centering
  \begin{tabularx}{\linewidth}{|X|}
    \hline
    When the Support strategy is effective:\\If you are being suspected by other players: Choose the Support strategy as it makes you less noticeable and helps you avoid suspicion.
If there are no other noticeable players: Choose the Support strategy to avoid drawing too much attention to yourself.
If potential Werewolves are not yet the focus of the discussion and have not been noticed: Acting inconspicuously is appropriate.\\

When the Attack strategy is effective:
If potential Werewolves are already being suspected: Take advantage of this momentum to strengthen criticism and solidify the suspicion.
If your teammates' credibility is reasonably established and the focus should shift to reducing the enemies' credibility: Actions to undermine opponents' credibility are effective.
If you want to steer the flow of discussion: Choose the Attack strategy when the intention is to take bold actions to change the situation.

- Based on the observations (conversation and latest possibility estimation), determine whether a Support or Attack strategy would be more effective.
\\
    \hline
  \end{tabularx}
   \caption{Adaptation Prompt of Villagers}
   \label{tab:vil_adap_Prompt}
\end{table}

\begin{table}
  \centering
  \begin{tabularx}{\linewidth}{|X|}
    \hline
When the Support strategy is effective:
If you are being suspected by other players: Choose the Support strategy as it makes you less noticeable and helps you avoid suspicion.
If there are no other noticeable players: Choose the Support strategy to avoid drawing too much attention to yourself.
If your teammate is being suspected: Choose the Support strategy and support your teammate's statements.\\

When the Attack strategy is effective:
If a potential opponent is suspected of being a Werewolf, take advantage of this momentum to strengthen the criticism and solidify the suspicion.
If your or your teammates' credibility is reasonably established, and the focus needs to shift to undermining the enemy's credibility, Actions to weaken the opponents' credibility are effective.
If you want to steer the flow of discussion, choose the Attack strategy when the intention is to take bold actions to change the situation.

- Based on the observations (conversation and latest possibility estimation), determine whether a Support or Attack strategy would be more effective.
\\
    \hline
  \end{tabularx}
  \caption{Adaptation Prompt of Werewolves}
  \label{tab:wolf_adap_Prompt}
\end{table}

\end{document}